\documentclass[manuscript, screen]{acmart}

\usepackage{algorithm}
\usepackage{algorithmic}

\AtBeginDocument{%
  }

\setcopyright{acmcopyright}
\copyrightyear{2023}
\acmYear{2023}
\acmDOI{XXXXXXX.XXXXXXX}

\acmJournal{TOMM}

\begin{document}

\title{Region-Aware Diffusion for Zero-shot Text-driven Image Editing}


\author{Nisha Huang}
\email{huangnisha2021@ia.ac.cn}
\orcid{0000-0002-1627-6584}
\affiliation{
  \institution{School of Artificial Intelligence, University of Chinese Academy of Sciences; National Laboratory of Pattern Recognition, Institute of Automation, Chinese Academy of Sciences}
  \country{China}
}

\author{Fan Tang}
\email{tfan.108@gmail.com}
\orcid{0000-0002-3975-2483}
\affiliation{%
  \institution{Institute of Computing Technology, Chinese Academy of Sciences}
  \country{China}
}

\author{Weiming Dong}
\email{weiming.dong@ia.ac.cn}
\orcid{0000-0001-6502-145X}
\affiliation{
    \institution{Institute of Automation, Chinese Academy of Sciences; School of Artificial Intelligence, University of Chinese Academy of Sciences}
    \country{China}
}

\author{Tong-Yee Lee}
\email{tonylee@mail.ncku.edu.tw}
\affiliation{
    \institution{National Cheng-Kung University}
    \country{Taiwan}
}

\author{Changsheng Xu}
\email{csxu@nlpr.ia.ac.cn}
\affiliation{
    \institution{Institute of Automation, Chinese Academy of Sciences; School of Artificial Intelligence, University of Chinese Academy of Sciences}
    \country{China}
}

\renewcommand{\shortauthors}{N. Huang et al.}

\begin{abstract}
    Image manipulation under the guidance of textual descriptions has recently received a broad range of attention.
    In this study, we focus on the regional editing of images with the guidance of given text prompts.
    Different from current mask-based image editing methods, we propose a novel region-aware diffusion model (RDM) for entity-level image editing, which could automatically locate the region of interest and replace it following given text prompts.
    To strike a balance between image fidelity and inference speed, we design the intensive diffusion pipeline by combing latent space diffusion and enhanced directional guidance.
    In addition, to preserve image content in non-edited regions, we introduce regional-aware entity editing to modify the region of interest and preserve the out-of-interest region.  
    We validate the proposed RDM beyond the baseline methods through extensive qualitative and quantitative experiments. 
    The results show that RDM outperforms the previous approaches in terms of visual quality, overall harmonization, non-editing region content preservation, and text-image semantic consistency.
    The codes are available at \url{https://github.com/haha-lisa/RDM-Region-Aware-Diffusion-Model.}
\end{abstract}



\begin{teaserfigure}
    \includegraphics[width=\textwidth]{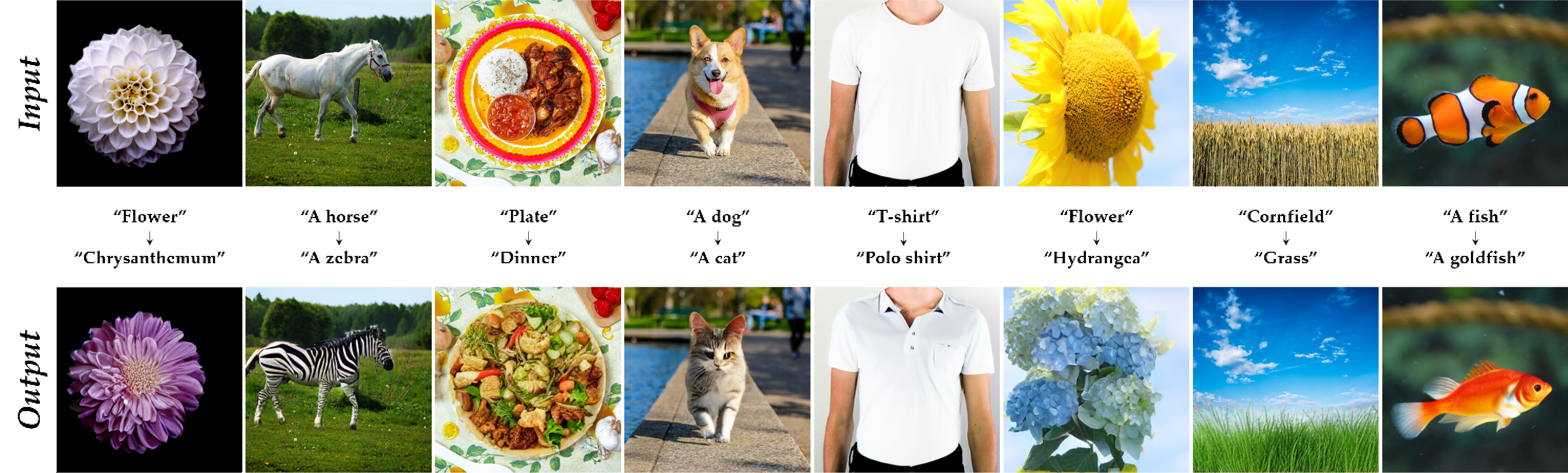}
    \caption{The results of the proposed region-aware diffusion model (RDM). The texts adhere to the phrase rule ``A → B'', indicating that RDM transforms entity A into entity B.}
    \label{fig:teaser}
\end{teaserfigure}

\maketitle

\section{Introduction}
In the actual world, image editing is highly sought-after.
However, image content editing software is not easy to get started with and is more suited to professionals.
Present image editing work is limited in terms of possible input images and editing operations.
In recent years, great advances in deep generative image models~\cite{vqgan, Guided-Diffusion, latentdiffusion} and visual language pre-training models~\cite{clip} have made text-based image generation and manipulation interfaces possible.

There are different principal branches of image generation, such as inpainting, image translation~\cite{choi2021ilvr}, style transfer~\cite{huang2017arbitrary}, and image manipulation~\cite{wang2022manitrans, nichol2021glide, ramesh2022hierarchical, avrahami2022blended, avrahami2022latentblended}. 
Computational approaches~\cite{openedit, bau2021paint, Flexit_2022_CVPR} for modifying the style and appearance of objects in natural photographs have made remarkable progress, allowing beginner users to accomplish a wide range of editing effects.
Nevertheless, it should be noted that prior text-based image manipulations work either did not allow for arbitrary text commands or image manipulation~\cite{mo2018instagan, bau2021paint, diffusionCLIP_2022_CVPR} or only allowed for modifications to the image's appearance properties or style~\cite{openedit, jiang2021language}.
Controlling the localization of modifications normally requires the user to draw a region to specify~\cite{avrahami2022blended, avrahami2022latentblended, bau2021paint}, which adds to the complexity of the operation.
In this study, we aim to eliminate all of the aforementioned constraints and restrictions to enable open image content modification with pure textual control utilizing cutting-edge image generation techniques.

Recently, there have been tremendous advances in multimodal deep learning that have opened the way for machines to achieve cross-modal communication and control.
One of the large-scale multimodal pre-training models that have received many applications is the Contrastive Language Image Pretraining (CLIP)~\cite{clip} model, which was pre-trained on $400$ million text-image samples.
In parallel, various new image synthesis methods~\cite{StyleCLIP, StyleGANNADA, CLIPstyler, diffusionCLIP_2022_CVPR, ramesh2022hierarchical, Huang2022MGAD} have highlighted the richness of the vast visual and linguistic realm encompassed by CLIP.
Nonetheless, manipulating existing items in arbitrary, actual pictures continues to remain tricky.
The present mainstream techniques combine CLIP with pre-trained GANs generators, however, the input picture domain is constrained.
There has recently been a tremendous amount of interest in diffusion models, which generate high-quality, diversified pictures.
Image manipulation at the pixel level, on the other hand, leads to extended generation times and excessive computer resource consumption.

After investigation, we found that applying the diffusion process in the latent space of pre-trained autoencoders~\cite{latentdiffusion} can speed up inference and reduce the consumption of computational resources.
Nevertheless, previous latent diffusion models still fall short in terms of generating image realism. 
To improve image realism and to enhance the consistency of the editing results with the guide text, we introduced classifier-free guidance~\cite{ho2021classifier} at each step of the diffusion.
Based on these, we develop a high-performance region-aware diffusion model (RDM) framework for zero-shot text-driven image editing.
Specifically, our method enables the editing of image content that satisfies an arbitrarily given text prompt (as shown in Fig.~\ref{fig:teaser}). 
For example, given an image of the dog and the positioning text: ``A dog'', our work can position the corresponding editing area. 
Then, based on the target text: ``A cat'', a high-quality, realistic, and varied image can finally be composed.

To enable the user to specify the area and objects to be modified and the objects to be created, simply and intuitively, we present a cross-modal entity calibration component. 
It can locate and adjust the text-relevant picture tokens given the positional textual guidance $t_1$ to correctly identify the entities for modification.
We observed that by feeding the input image into the encoder together with the mask, the content outside of the mask appeared to transform unexpectedly during the diffusion process for image modification.
To retain extraneous content to a greater extent, we perform further diffusion at each diffusion step where it blends the clip-guided diffusion result with the corresponding noisy version of the input image.
In addition, we build relevant loss functions that protect non-editing domains to constrain the generative process.
We incorporated the clip gradient into the classifier-free guidance to make the edited results more favorable to humans and to make the content generated in the edit area more consistent with the semantic content of the target textual guidance $t_2$.
Overall, RDM implements enhanced directional guidance in the latent space to generate rapid, high-quality, realistic, and text-compliant editing results.

Quantitative and qualitative comparisons with previous approaches reveal that our method can better manipulate the entities of an image through text while leaving the background region unaffected. 
As shown in Figs.~\ref{fig:teaser},~\ref{fig:moreresults2} and \ref{fig:moreresults1}, 
RDM is capable of producing realistic and high-quality outcomes in terms of object content change when guided by various image inputs and text descriptions.
The main contributions of this work are summarized as follows:
\begin{itemize}
\item We propose a novel region-aware diffusion model (RDM), an entity-level zero-shot text-driven image editing framework based on the intensive diffusion model.
\item We introduced spatial location masks into each step of the diffusion sampling and created non-editing region-preserving loss functions to obtain edited results without stitching traces and well-preserved unedited regions.
\item We manipulate the diffusion step in latent space and embed enhanced directional guidance structures to enhance image realism and improve the consistency of the control text with the editing result.
\item The quantitative and qualitative experimental results show that RDM outperforms baseline methods in terms of quality, veracity, and diversity in text-guided image editing, and achieves superior results.
\end{itemize}
\section{Related Work}
In this section, we review the existing works on text-guided image manipulation and diffusion models, which motivates us to design and implement our application.

\begin{figure*}[tbp]
    \centering
    \includegraphics[width=\linewidth]{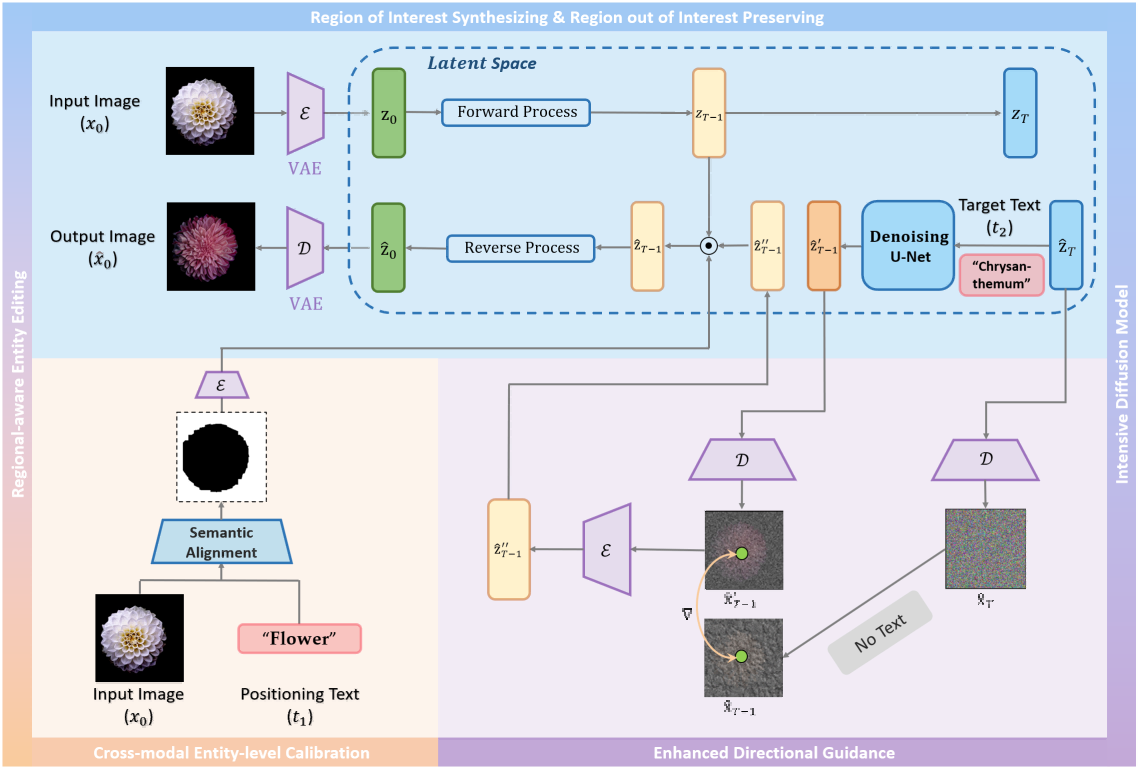}
    \caption{The overall framework of our method for zero-shot text-driven image editing.}
    \label{fig:pipeline}
\end{figure*}

\subsection{Text-guided Image Manipulation}
Synthesizing an image based on a text description is an ambitious problem that has advanced tremendously in recent years. 
Initial RNN-based works~\cite{mansimov2015generating} were surpassed by generative adversarial approaches. 
There have been seminal works based on conditional GANs in image editing~\cite{dong2017semantic, li2020manigan,nam2018text}. 
``Paint By Word''~\cite{bau2021paint} firstly addressed the problem of zero-shot semantic image painting using CLIP~\cite{clip} in combination with StyleGAN2~\cite{karras2020analyzing} and BigGAN~\cite{brock2018large}.
It can only alter the appearance of a picture, such as its color and texture, but it cannot generate new entities.
ManiGAN~\cite{li2020manigan} semantically edits parts of an image matching a given text that describes certain attributes and preserving the contents irrelevant to the text.
However, the expressiveness of the text is restricted by such multimodal GAN-based approaches.
Both ``Paint By Word''~\cite{bau2021paint} and ManiGAN~\cite{li2020manigan} are restricted to specific image domains and are not applicable to open natural images.

SDG~\cite{liu2021more} and DiffusionCLIP~\cite{diffusionCLIP_2022_CVPR} are proposed to utilize a diffusion model in order to perform global text-guided image manipulations. 
GLIDE~\cite{nichol2021glide} and DALL·E 2~\cite{ramesh2022hierarchical} focus on text-driven open domain image synthesis, as well as local image editing. 
GLIDE fine-tunes its text-to-image synthesis model for image inpainting. 
DALL·E 2 performs inpainting results while lacking discussion in the paper.
Both of them are implemented with the idea of integrating image generators and joint text-image encoders into their architectures. 
They all contain pre-trained models with large-scale datasets of numerous text-image pairs, while neither of them has released their complete models.
Later, Blended Diffusion~\cite{avrahami2022blended} and Latent Blended Diffusion~\cite{avrahami2022latentblended} were proposed as the solution for local text-guided editing of real generic images. 
However, these methods require the user to draw the extra mask manually from which the image is edited, without a precise and automatic editing area.

\begin{figure*}[thbp]
    \centering
    \includegraphics[width=\linewidth]{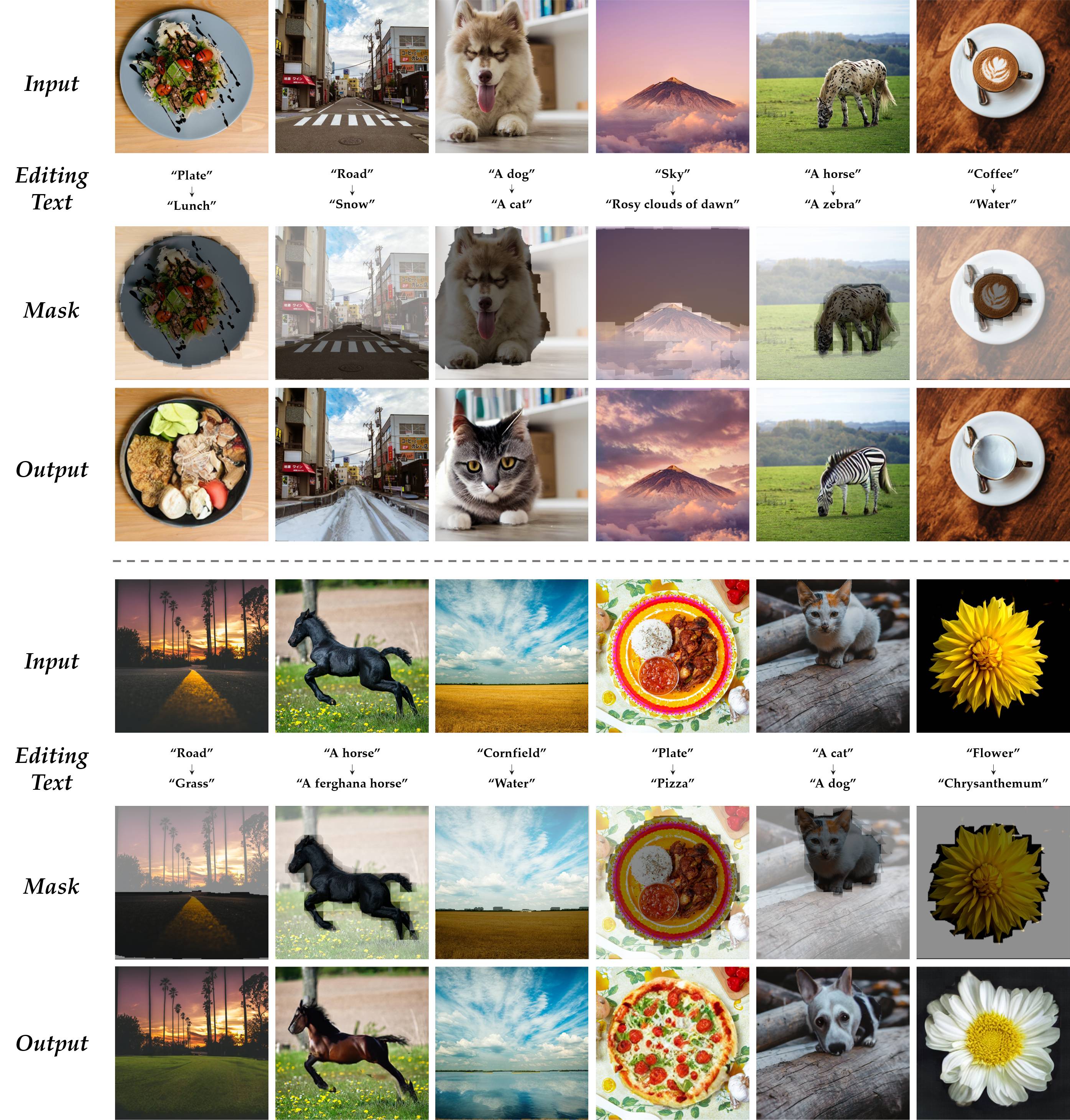}
    \caption{More manipulation results by RDM. Across rows: different inputs; across columns: diverse editing masks and results.}
    \label{fig:moreresults2}
\end{figure*}

\subsection{Diffusion Models}
Diffusion models, also known as score-based generative models, are a strong family of generative models that have recently evolved.
This fresh idea on the subject of image generation was proposed by Sohl-Dickstein et al.~\cite{sohl2015deep}. 
Current works~\cite{Guided-Diffusion, AlexiaJolicoeurMartineau2021AdversarialSM, DDIM} demonstrate astonishing results in high-fidelity image generation, often even outperforming generative adversarial networks. 
Importantly, \cite{nichol2021glide, ramesh2022hierarchical, saharia2022photorealistic} additionally offer strong sample diversity and faithful mode coverage of the learned data distribution. 
As a result, diffusion models are ideal for learning models from complicated and varied data.

Specifically, diffusion models consist of one forward process and one reverse process.  
The forward diffusion process maps data to noise by gradually perturbing the input data.
The reverse process performs iterative denoising from pure random noise. 
The diffusion models are used to generate data by simply passing randomly sampled noise through the learned denoising process.
Diffusion models have already been utilized in many successful applications, such as image generation~\cite{Guided-Diffusion, AlexiaJolicoeurMartineau2021AdversarialSM,  nichol2021glide, ramesh2022hierarchical, saharia2022photorealistic, DDIM}, image segmentation~\cite{baranchuk2021labelefficient}, image-to-image translation~\cite{choi2021ilvr}, superresolution~\cite{kawar2022denoising, latentdiffusion}, and image editing~\cite{avrahami2022latentblended,avrahami2022blended,nichol2021glide, ramesh2022hierarchical}.
Text2LIVE~\cite{bar2022text2live} applies the text to edit the appearance of existing objects.
It concentrates on generating an edit layer composited over the original input, rather than removing or replacing objects of the input image, as we do.

Even though the approaches described above produce cutting-edge outcomes for picture data generation, one disadvantage of diffusion models is the sluggish reverse denoising process.
In addition, traditional diffusion models operate in pixel space leading to consuming a lot of memory.
Latent diffusion models (LDMs)~\cite{latentdiffusion} have been proposed to expedite the sampling process and reduce computational requirements compared to pixel-based diffusion models. 
LDMs are trained to build latent visual representations and to perform the diffusion process across a lower-dimensional latent space.
\cite{latentdiffusion} shows that it has achieved a new state-of-the-art and highly competitive performance on various computer vision tasks.
However, there is still a need to improve performance in terms of image fidelity and text-image semantic consistency.
Therefore, our RDM is designed to retain the benefits of LDM speed while taking into account image quality and text-image alignment.

\section{Method}
\subsection{Overview}


The proposed RDM is a framework for solving entity-level zero-shot text-driven image editing tasks, as depicted in Fig.~\ref{fig:pipeline}. Our goal is to implement editing of the input image $x_0$ through the control by a pair of text prompts ($t_1,~t_2$). 
The positioning text $t_1$ is used to position the edited entity, and the target text $t_2$ is used to generate the new entity.
In Section~\ref{subsec: reinforced diffusion}, we illustrate the concrete composition of the intensive diffusion model. 
In Section~\ref{subsec: Semantic Alignment}, we explain how regional-aware entity editing can be achieved through text.

\begin{figure}[thbp]
    \centering
    \includegraphics[width=0.8\linewidth]{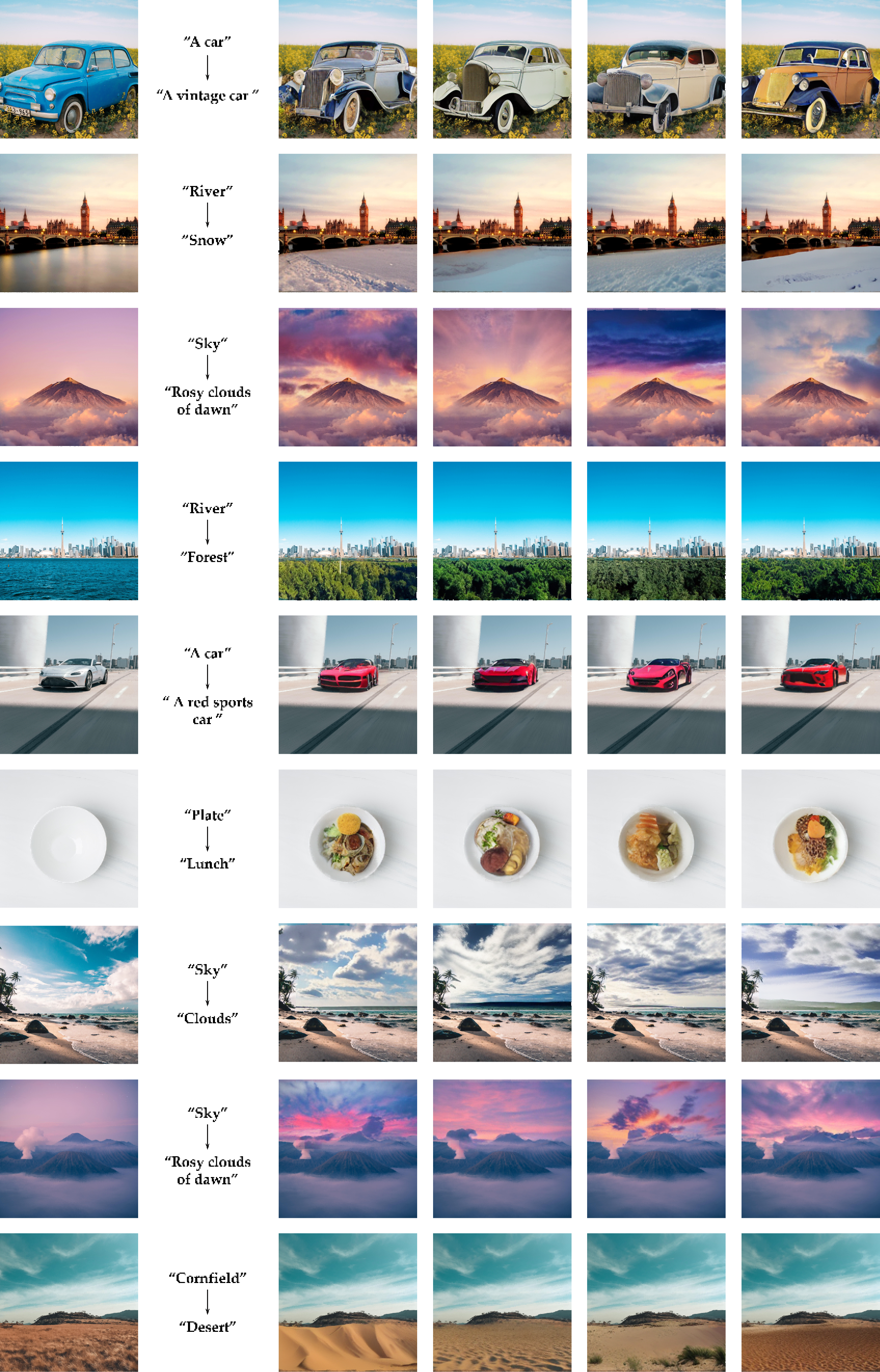}
    \caption{More manipulation results by RDM.}
    \label{fig:moreresults1}
\end{figure}
\subsection{Intensive Diffusion Model}
\label{subsec: reinforced diffusion}

The diffusion model~\cite{sohl2015deep} is a generator that can be used to generate images.
The diffusion process is divided into a forward process, which adds random noise to the input image $x_0$, and a backward process, which removes the noise and generates the image $\hat{x}_0$.
Unlike traditional diffusion models~\cite{Guided-Diffusion, DDPM}, we do not perform the diffusion process at the pixel level. Some recent works~\cite{avrahami2022latentblended, latentdiffusion} have demonstrated that performing the diffusion process in the latent space can reduce computational consumption and speed up the sampling process.
The denoising UNet learns to remove the noise, and after $T$ steps of noise removal, generates the output image $\hat{x}_0$.
However, there is damage to image generation quality and text-image consistency by performing the diffusion process in the latent space. 
To improve this issue, we further introduce a component of enhanced directional guidance.



\paragraph{Latent Representations.}
As mentioned above, we perform a diffusion step in the latent space~\cite{latentdiffusion} to reduce complexity and provide efficient image processing.
An autoencoder VAE~\cite{vae} is used to accomplish perceptual picture compression.
The diffusion model directly operates on the lower-dimensional latent space, taking advantage of image-specific inductive biases. 
This allows the underlying autoencoder to be constructed primarily from two-dimensional convolutional layers and uses a re-weighting bound to further focus the target on the perceptually most important bits, which are denoted as:
\begin{equation}
L_{L D M}:=\mathbb{E}_{\mathcal{E}(x), \epsilon \sim \mathcal{N}(0,1), t}\left[\left\|\epsilon-\epsilon_\theta\left(z_t, t\right)\right\|_2^2\right].
\end{equation}
Our model's determination $\epsilon_\theta(z_t, t)$ is implemented as a
time-conditional UNet~\cite{unet}. 
Given that the forward process is fixed, $z_t$ can be conveniently acquired from $\mathcal{E}$ during training, and $\mathcal{D}$ can decode samples from $p(z)$ to pixel space.

\paragraph{Enhanced Directional Guidance.} To reinforce the editing direction of the source region to follow the target text, we attempt to modify a classifier-free guidance~\cite{ho2021classifier} to strengthen cross-modal guidance.
It is a strategy for guiding diffusion models without necessitating the training of a separate classifier model.  
Generally, classifier-free guidance offers two benefits. 
For starters, rather than relying on the knowledge of a separate (and perhaps smaller) categorization model, it allows a single model to leverage its experience while guiding.
Second, it simplifies directing when conditioned on information that is difficult to predict using a classifier.

In order to provide classifier-free guidance, the tag $y$ in a class-conditional diffusion model $\epsilon_{\theta}\left(x_{t} \mid y\right)$ is replaced with a null tag $\emptyset$ throughout the training process.
The output of the model is further extended in the direction of $\epsilon_{\theta}\left(x_{t} \mid y\right)$ and away from $\epsilon_{\theta}\left(x_{t} \mid \emptyset\right)$ during sampling:
\begin{equation}
\hat{\epsilon}_{\theta}\left(x_{t} \mid y\right)=\epsilon_{\theta}\left(x_{t} \mid \emptyset\right)+s \cdot\left(\epsilon_{\theta}\left(x_{t} \mid y\right)-\epsilon_{\theta}\left(x_{t} \mid \emptyset\right)\right).
\end{equation}
The recommended guidance scale is $s = 5$. This equation was inspired by the classifier,
\begin{equation}
p^{i}\left(y \mid x_{t}\right) \propto \frac{p\left(x_{t} \mid y\right)}{p\left(x_{t}\right)}.
\end{equation}
The function of the true scores is used to represent the gradient $\epsilon^{*}$,
\begin{equation}
\begin{aligned}
\nabla_{x_{t}} \log p^{i}\left(x_{t} \mid y\right)  \propto \nabla_{x_{t}} \log p\left(x_{t} \mid y\right)-\nabla_{x_{t}} \log p\left(x_{t}\right),  \propto \epsilon^{*}\left(x_{t} \mid y\right)-\epsilon^{*}\left(x_{t}\right).
\end{aligned}
\end{equation}
The modified prediction $\hat{\epsilon}$ is subsequently employed to guide us toward the target text prompts $t_2$, as demonstrated in Algorithm~\ref{alg1}:
\begin{equation}
\hat{\epsilon}_{\theta}\left(x_{t} \mid t_{2}\right)=\epsilon_{\theta}\left(x_{t} \mid \emptyset\right)+s \cdot\left(\epsilon_{\theta}\left(x_{t} \mid t_{2}\right)-\epsilon_{\theta}\left(x_{t} \mid \emptyset\right)\right).
\end{equation}

\subsection{Regional-aware Entity Editing}
\label{subsec: Semantic Alignment}
\paragraph{Cross-modal Entity-level Calibration.}
To generate a binary segmentation mask $m$ based on the localized text $t_1$, we design a cross-modal entity-level calibration module, consisting of a pre-trained CLIP model and a thin conditional segmentation layer (decoder).
First, the positioning text $t_1$ is fed into the CLIP text transformer to obtain the conditional vector. Motivated by feature-wise transformations~\cite{film,clipseg}, the conditional vector is used to modulate the input activation of the decoder. This enables the decoder to associate the activation within CLIP with the output segmentation and to inform the decoder about the segmentation's target.
The input image $x_{0}$ is passed through the CLIP visual transformer to get $\mathbb{R}^{W\times H\times 3}$. 
Afterward, the activations extracted at layers $S = [3,7,9]$ are added to the decoder internal activations at the embedding size $F = 64$ before each transformer block.
Besides, CLIP ViT-B/16 is used with token patch size: $P = 16$. 
The decoder generates the binary segmentation by applying a linear projection on the tokens of its transformer (last layer):
\begin{equation}
\mathbb{R}^{\left(1+\frac{W}{P} \times \frac{H}{P}\right) \times F} \mapsto \mathbb{R}^{W \times H}.
\end{equation}
To associate CLIP's capabilities with segmentation results, a generic binary prediction setting is employed.
We threshold the binary segmentation for spatial mask $m$, with a threshold $K$ ranging from 0 to 255, which we usually take as 150.

\paragraph{Region of Interest Synthesizing.}
To make the region of interest could be edited according to the text prompt, we leverage a pre-trained ViT-L/14 CLIP~\cite{clip} model for text-driven image content manipulation.
The cosine distance between the CLIP embedding of the denoised image $\hat{x}_{t}$ during diffusion and the CLIP embedding of the text prompt $t_2$ may be used to specify the CLIP-based loss, or $\mathcal{L}_{C L I P}$.
Target textual prompt $t_{2}$ is embedded into the embedding space, which is defined as $E_{L}$.
And a time-dependent image encoder for noisy images is referred to as $E_{I}$.
We define the language guidance function using the cosine distance, which measures how similar the embeddings $E_{I}$ and $E_{L}$ are to one another.
The text guidance function can be defined as:
\begin{equation}
\mathcal{L}_{C L I P}\left(\hat{x}_{t}, t_{2}, m\right)=E_{I}\left(\hat{x}_t \odot m\right) \cdot E_{L}(t_{2}).
\end{equation}

The aforementioned process is not subject to any extra non-editing region restrictions.
Despite being assessed inside the region that is being edited, $\mathcal{L}_{C L I P}$ also affects non-editing regions.
We provide the equivalent approach below to deal with this problem.

\paragraph{Region out of Interest Preserving.}
Non-editing region preserving (NERP) is not present in the aforementioned procedure, which starts with isotropic Gaussian noise.
As a result, even though $\mathcal{L}_{C L I P}$ is assessed inside the masked zone, it still has an impact on the whole image.
To ameliorate this problem, we encode the mask $m$ into the latent space to get $m_{\text {latent}}$, and blend it into the diffusion process as follows.
The latent for the subsequent latent diffusion step is produced by blending the two results using the resized mask, i.e. $\hat{z}_{t}^{\prime \prime} \odot m_{\text {latent}}+{z}_{t} \odot\left(1-m_{\text {latent}}\right)$.
As shown in Fig.~\ref{fig:pipeline}, where $z^{\prime \prime}$ represents the result generated by a latent diffusion followed by enhanced directional guidance in the reverse process.
And $z_{nd}$ is the result of superimposing the corresponding noise on the input image in the forward process.
At each denoising step, the entire latent is modified, but the subsequent blending enforces the parts outside $m_{\text {latent}}$ to remain the same.
In this stage, the backdrop is tightly preserved by replacing the whole area outside the mask with the comparable region from the input image.
The subsequent latent denoising process ensures coherence even though the resultant blended latent is not always coherent.
Following the completion of the latent diffusion process, we decode the resulting latent to the output image using the decoder $\mathcal{D}(z)$, as demonstrated in Algorithm~\ref{alg1}.

Besides, a non-editing region preserving loss $\mathcal{L}_{N E R P}$ is applied to direct the diffusion outside the mask to direct the surrounding area towards the input image:
\begin{flalign}
& \mathcal{L}_{N E R P}\left(x_{0}, \hat{x}_t, m\right)=d\left(x_{0} \odot(1-m), \hat{x}_t \odot(1-m)\right), \\
& d\left(a, b\right)=\lambda_{1}\left(LPIPS\left(a, b\right)\right)+\lambda_{2}\left(MSE\left(a, b\right)\right),\\
& a=x_{0} \odot(1-m), b=\hat{x}_t \odot(1-m).
\end{flalign}
where $LPIPS$ is the learned perceptual image patch similarity measure and $MSE$ is the $\operatorname{L_2}$ norm of the pixel-wise difference between the images.
$\lambda_{1}$ and $\lambda_{1}$ are all set to $0.5$.

\begin{algorithm}[htb]  
  \caption{Text guided hybrid diffusion sampling, given \\ a latent diffusion model $(\mu_{\theta}(z_{t}), \Sigma_{\theta}(z_{t}))$}  
  \label{alg1}  
  \begin{algorithmic}[1]  
    \REQUIRE The input image $x_0$, text guidance $t_{2}$, gradient scale $s$, diffusion steps $T$.
		\ENSURE generated image ${x}_{0}$ according to text guidance $t_{2}$.
		\STATE $t = T$
		\STATE $z_{0} = \mathcal{E}(x_{0})$
		\STATE $z_{t} \leftarrow$ sample from $\mathcal{N}(0, \mathbf{I})$
		\STATE $\hat{z}_{T} = {z}_{T}$
		\REPEAT
		\STATE $t - 1 \gets t$ 
	        \STATE $\mu, \Sigma \leftarrow \mu_{\theta}(\hat{z}_{t}), \Sigma_{\theta}(\hat{z}_{t})$
            \STATE $\hat{z}_{t-1}^{\prime} =$ denoise $(\hat{z}_{t}, t_2, t)$
            \STATE $\hat{x}_{t-1}^{\prime} = \mathcal{D}(\hat{z}_{t-1}^{\prime})$
            \STATE ${\hat\epsilon}_{\theta}( \hat{x}_{t-1} \mid t_{2}) \leftarrow  ( 1-s )\cdot {\epsilon _{\theta }} ( \hat{x} _{t-1}\mid \emptyset )+s \cdot\epsilon_{\theta}( \hat{x}_{t-1}^{\prime} \mid t_{2}) $
            \STATE $\mathcal{L} \gets \mathcal{L}_{C L I P}( \hat{x}_{t-1}^{\prime}, t_2, m) + \mathcal{L}_{N E R P}(x_{0}, \hat{x}_{t-1}^{\prime}, m)$
            \STATE $\hat{z}_{t-1}^{\prime \prime} \leftarrow$ sample from $\mathcal{N}(\mu+s\Sigma\nabla_{\hat{x}_{t-1}}\mathcal{L})$
            \STATE $\hat{z}_{t-1} =  \hat{z}_{t-1}^{\prime \prime}\odot m_{\text {latent}}+ z_{t-1} \odot(1-m_{\text {latent}})$
		\UNTIL {$t < 0$} 
		\STATE $\hat{x}_0 = \mathcal{D}(\hat{z}_{0})$
  \end{algorithmic}  
\end{algorithm} 
\begin{figure*}[thbp]
    \centering
    \vspace{-0.8cm} 
    \includegraphics[width=0.83 \linewidth]{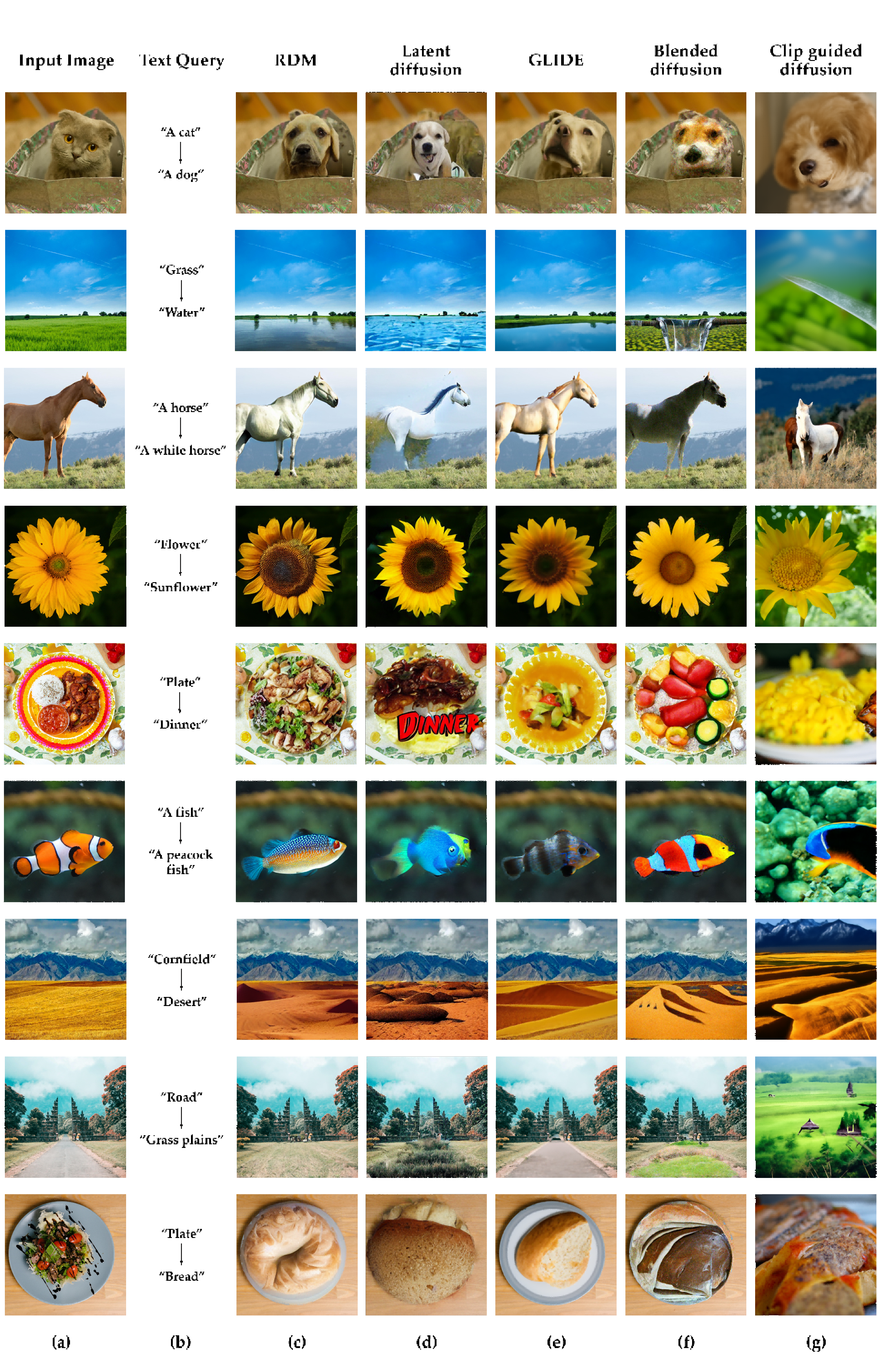}
    \vspace{-0.5cm} 
    \caption{Comparison with SOTAs including latent diffusion, GLIDE, blended diffusion, and CLIP-guided diffusion.}
    \label{fig:comparison}
\end{figure*}
\begin{figure*}[btp]
    \centering
    \includegraphics[width=\linewidth]{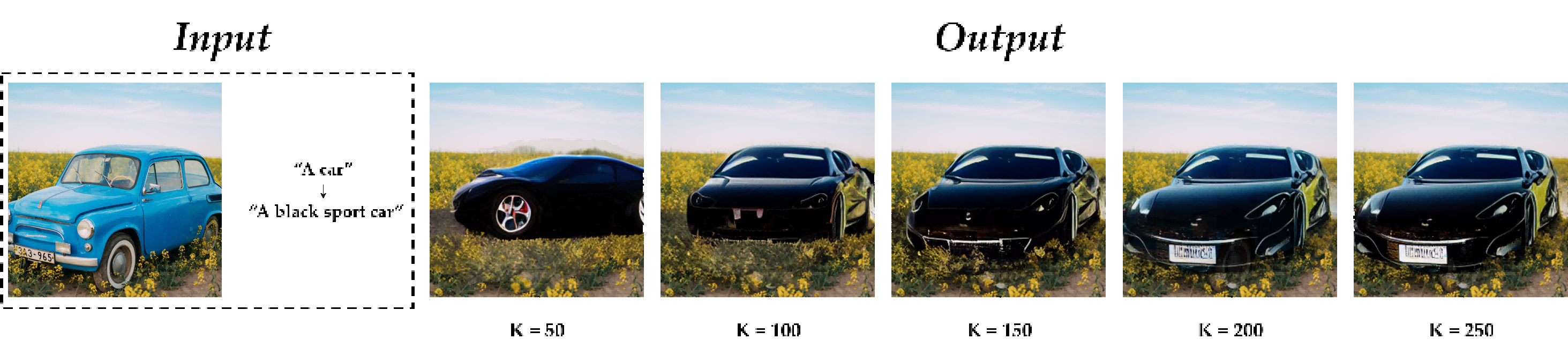}
    \caption{Impact of mask threshold on the manipulation results.}
    \label{fig:ablation_mask_threshold}
\end{figure*}
\begin{figure}[btp]
    \centering
    \includegraphics[width=\linewidth]{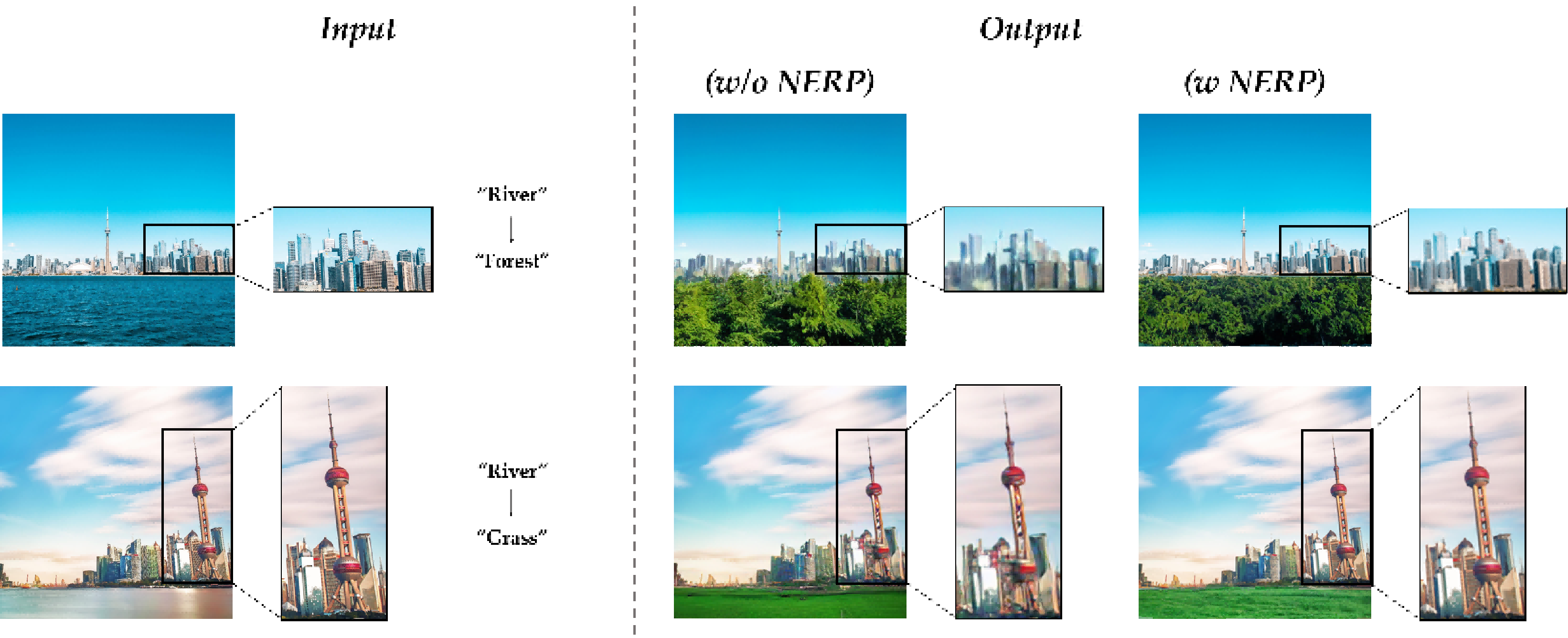}
    \caption{Qualitative comparison of our RMD with or without non-editing region preserving component.}
    \label{fig:ablation_nerp}
\end{figure}
\begin{figure*}[btp]
    \centering
    \includegraphics[width=\linewidth]{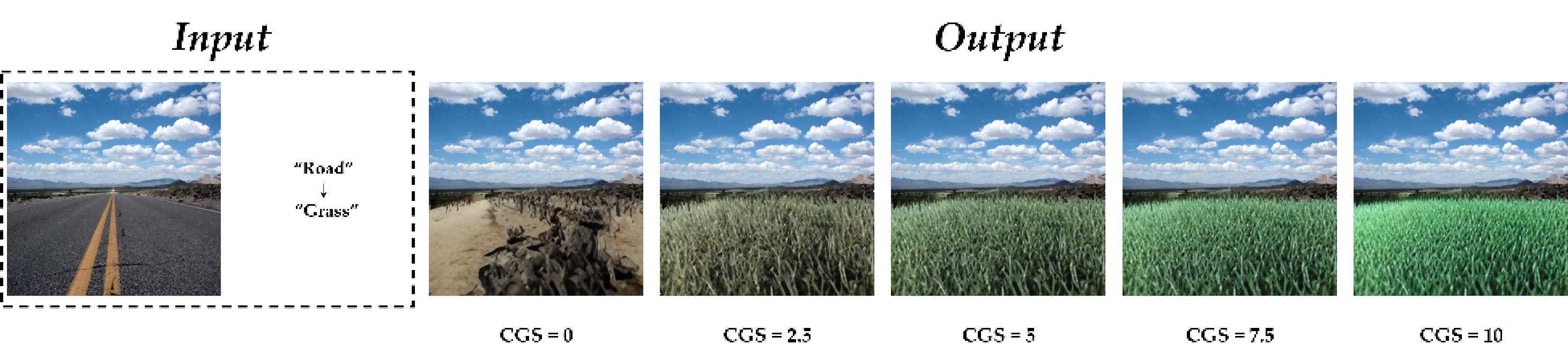}
    \caption{Effect of classifier-free guidance scale (CGS) on the generated results.}
    \label{fig:ablation_cgs}
\end{figure*}
\section{Experiments}
\subsection{Implementation Details}
For the diffusion model, we used a pre-trained latent diffusion model~\cite{latentdiffusion} of resolution $256 \times 256$, which has $1.45$ billion parameters trained on the LAION-400M~\cite{schuhmann2021laion} database. 
For the CLIP model, we used ViT-L/14 released by OpenAI for the Vision Transformer~\cite{clip}.
The output size of RDM is $256 \times 256$.
For sampling, we set $\lambda_{1}$, $\lambda_{2}$ and clip guidance scale to $0.5$, $0.5$ and $150$, respectively.
To ensure the quality of the results and to maintain the consistency of the parameters, the diffusion step and the time step used for the experiments in this work are both set to $100$. 
It takes three seconds to generate a $256 \times 256$ image on a single GeForce RTX $3090$ GPU by RDM, which is comparable to latent diffusion (three seconds) and surpassing most diffusion models ($15$ seconds, $27$ seconds, and $3$ minutes respectively for GLIDE, blended diffusion, and clip-guided diffusion).

\subsection{Qualitative Evaluation}
We tested our approach on a variety of real-world images and edited texts. 
The images were sourced from the web and contained a variety of object categories, including animals, food, landscapes, and others. 
Examples of the input and our results can be seen in Figs.~\ref{fig:teaser},~\ref{fig:moreresults2} and \ref{fig:moreresults1}.
Fig.~\ref{fig:moreresults2} shows the segmentation masks generated by our method guided by $t_1$ for arbitrary images.
The results show that our method does not produce obvious artificial editing traces by using different images and text prompts as inputs.
As shown in the flower in Fig.~\ref{fig:teaser}, the transition between the flower and the background is natural and the petals are generated with natural texture.
For food manipulation, it is possible to see each piece of food. 
In the editing of animals, RDM can compose the new animal very well even though the animal being edited has multiple complex poses. 
As shown in Fig.~\ref{fig:moreresults1}, RDM can obtain a variety of different results for the same image and text input. 
In a nutshell, our method successfully applies text control to the editing of image entity content in high quality and diversity.

\subsection{Comparison with the State-of-the-arts} 
In this section, we compare RDM with SOTA text-driven image editing methods including Latent diffusion~\cite{latentdiffusioninpaint}, GLIDE~\cite{nichol2021glide}, blended diffusion~\cite{avrahami2022blended} and CLIP-guided diffusion~\cite{clip-guideddiffusion}.
Fig.~\ref{fig:comparison} shows comparisons to baselines of real-world images. 
The main differences between our approach and these methods are as follows.
Latent diffusion~\cite{latentdiffusioninpaint}, GLIDE~\cite{nichol2021glide}, and blended diffusion~\cite{avrahami2022blended} require the user to provide a mask for the area to be edited. 
Their out-of-interest regions are not involved in diffusion, so there are no content-preserving issues. 
CLIP-guided diffusion~\cite{clip-guideddiffusion} cannot be modified for local areas of the image. 

In this comparison, the masks required for these methods are provided by masks generated by RDM's cross-modal entity-level calibration component. 
As can be seen from the results (in column (d) of Fig.~\ref{fig:comparison}) of latent diffusion~\cite{latentdiffusion}, even though a strict edit mask is provided, the new content generated does not match the area of the mask and always generates new content that is smaller than the masked area. 
The results of GLIDE~\cite{nichol2021glide} (in column (e) of Fig.~\ref{fig:comparison}) lack details; for example, the petals of the sunflower are very smooth; the skin of the horse has no texture, and the hair on the horse's back is lacking; the ``grass plains'' generation fails and GLIDE does not understand the content of this text prompt well. 
The images (in column (f) of Fig.~\ref{fig:comparison}) produced by blended diffusion~\cite{avrahami2022blended} lack realism and are artificial. 
The results (in column (g) of Fig.~\ref{fig:comparison}) of CLIP-guided diffusion~\cite{clip-guideddiffusion} do not preserve the content of the unmodified areas of the image and tend to over-vignette.

\begin{table*}[htbp]
  \centering
    \caption{Quantitative comparisons for image manipulation. We compute the average CLIP score, SFID score and image harmonization (IH) score to measure visual quality, text consistency, and image harmonization. The best results are highlighted in \textbf{bold} while the second best results are marked with an \underline{underline}}
  \label{tab:quantity}%
    \begin{tabular}{cccccc}
    \toprule
                & RDM   & Latent diffusion & GLIDE       & Blended diffusion & \multicolumn{1}{p{8.875em}}{CLIP guided diffusion} \\
    \midrule
    CLIP score ↑ & \textbf{0.849 } & 0.824      & \underline{0.845}      & 0.822      & 0.843  \\
    SFID score ↓   & \underline{6.54}      & 9.29     & \textbf{5.88 }  & 17.37      & 23.42  \\
    IH score ↓   & \textbf{20.7 } & 22.0       & \underline{21.8}       & 23.1       & / \\
    \bottomrule
    \end{tabular}%
  \vspace{+0.4cm} 

\end{table*}%
\begin{figure}[tbp]
    \centering
    \includegraphics[width=8.5cm]{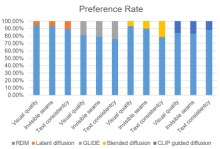}
    \caption{User study results. Each number represents the percentage of votes received by the other models' outcomes as compared to our results.}
    \label{fig:userstudy}
\end{figure}

\subsection{Quantitative Evaluation}

\paragraph{CLIP score.}
To assess the semantic alignment of the text descriptions and modified images, we compute the CLIP score, which is the cosine similarity between their embeddings derived with CLIP encoders.
Because we utilize the ViT-B/14 CLIP model during the inference process, for a fair comparison, we compute the CLIP score using the ViT-B/32 CLIP~\cite{clip} model.
A higher CLIP score suggests that the input texts and altered images are semantically aligned.
The results (the 1$^{st}$ row of Table~\ref{tab:quantity}) show that RDM outperforms baseline models, which indicates a superior in terms of text image consistency.


\paragraph{SFID}
To assess the quality of manipulated images, we employ the SFID~\cite{kim2020simplified}, a simplified FID that avoids the numerical instability associated with a limited number of sample feature distributions.
We calculated the SFID score between the different methods to obtain the 2$^{nd}$ row of Table~\ref{tab:quantity}.
Intuitively, the lower the SFID scores, the higher quality of the manipulated images on the COCO~\cite{coco} dataset.
The findings reveal that GLIDE has the highest SFID score and RDM is the second-best.

\paragraph{Harmonization score.}
To evaluate the degree of harmonization between the edited and unedited parts, we used DoveNet~\cite{DoveNet2020} as a quantitative evaluation method.
We utilize DoveNet to generate harmonized images and calculate the PSNR values between the harmonized images and manipulated images.
Image harmonization (IH) scores are shown in the 3$^{rd}$ row of Table~\ref{tab:quantity}.
The lower the IH score, the more synchronized the edited and unedited parts are.
Our approach outperforms the baseline model by a margin, achieving the highest harmonization score.
Our incorporation of the mask into the diffusion process resulted in better performance in terms of consistency between edited and unedited regions.

\paragraph{User study.}
Next, we conducted a user perception evaluation.
Participants were asked to choose: which image produced a higher quality image, which image turned out to be more harmonious and had less visible editing marks (seams), and which image editing turned out to be more in line with the text content.
The judgments of $70$ participants were collected across $36$ image-text combinations and gathered $2520$ votes.
Each comparison was made without revealing which image was by which method.
We have included in Fig.~\ref{fig:userstudy} report the percentage of votes in favor of the RDM model.
It follows that our method is capable of generating the kind of image editing results that humans prefer.

\subsection{Ablation Study}
\paragraph{Effects of cross-modal entity-level calibration component.}
To investigate the impact of the cross-modal entity-level calibration component on the quality and semantic content of the generated images, we tested image editing at different thresholds $K$. Fig.~\ref{fig:ablation_mask_threshold}  shows that when the value of $K$ is set small, the segmented editable scene is also smaller than the area occupied by the vehicles in the input image to varying degrees. The orientation of the generated vehicles does not match the orientation of the vehicles in the input image, i.e., there is a deviation in semantic consistency. As the value of $K$ increases, the editable area increases, and the orientation of the vehicles tends to be the same, but after a certain point, the image editing results do not change significantly.

\begin{table}
  \centering
  \caption{Quantitative comparison of our RMD with or without the NERP component.}
  \label{tab:ablation}%
    \begin{tabular}{lrr}
    \toprule
                & \multicolumn{1}{l}{w NERP} & \multicolumn{1}{l}{w/o NERP} \\
    \midrule
    LPIPS ↓      & 0.039       & 0.143 \\
    \bottomrule
    \end{tabular}%
\end{table}%


\begin{figure}[tbp]
    \centering
    \includegraphics[width=8.5cm]{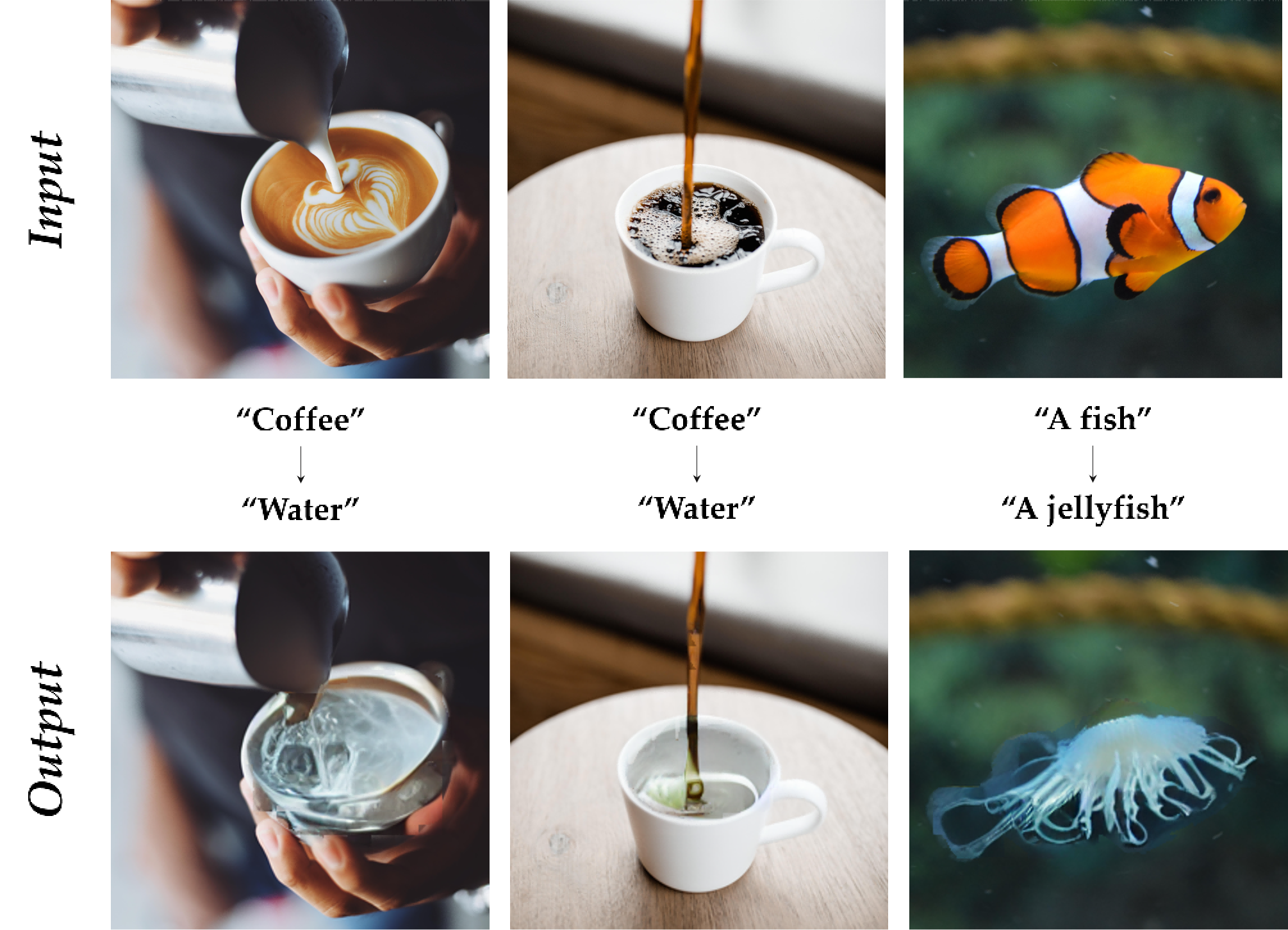}
    \caption{Examples of failure cases are given source images.}
    \label{fig:failurecase}
\end{figure}
\paragraph{Effects of the non-editing region preserving component.}
We perform a qualitative and quantitative comparison of the editing results with and without the non-editing region preserving (NERP) component.
As can be seen from Fig.~\ref{fig:ablation_nerp}, without the NERP component, the buildings in the image become more distorted with unusual artifacts and textures, leading to an overall loss of image quality.
The Learned Perceptual Image Patch Similarity (LPIPS)~\cite{LPIPS} is widely used to measure the difference between two images. We use LPIPS to measure the degree of retention of non-interest regions before and after editing. Lower LPIPS values mean better content preservation.
As can be seen from Table~\ref{tab:ablation}, this component enables us to help achieve image content preservation in the out-of-interest region of the input image.
Thus, the NERP component greatly improves the abnormal distortion and corruption of the image content outside the editing region.

\paragraph{Impact of classifier-free guidance scale.}
The classifier-free guidance is a technique that greatly improves the generation quality and image text alignment in a text conditional diffusion model. We investigate the effect of the classifier-free guidance on the results in Fig.~\ref{fig:ablation_cgs}.
We find that the results obtained without the classifier-free guidance are not competitive. Better results are generated when $s > 2.5$. In our experiments, we use the default guidance value of $s = 5$.



\subsection{Failed case.}
We have observed through some experiments that CLIP has a significant preference for particular solutions for various editors. 
As shown in Fig.~\ref{fig:failurecase}, given a picture of a cup with coffee, we wanted to implement a ``coffee'' to ``water'' image edit. 
The result shows that the liquid in the cup is successfully turned into water.
However, the text ``water'' is closely associated with a transparent cup, so it is possible that the cup could also be turned into a glass. 
As shown in the second column of Fig.~\ref{fig:failurecase} the coffee that is in the air being injected into the cup fails to successfully turn into water.
Lastly, we also notice that the editing results are not satisfactory when the editing target differs significantly in shape from the source object. As illustrated in the third column of Fig.9, the execution of "A fish" to "A jellyfish" produced additional boundary artifacts.
It should be noted that our method is more suited to generating a new entity rather than modifying the properties of the original entity.


\section{Conclusion and Future Work}
This paper investigates for the first time a new problem setting - the editing of the content of specified entities in images, guided by arbitrary text. 
Solving this task requires control over the positioning of the edits, the quality, and fidelity of the edited and unedited content, the consistency of text guidance and image manipulation, etc. To address these issues, we propose a new framework, a region-aware diffusion model with semantic alignment and generation capabilities, for manipulating images at the entity level. We provide a new tool for users to modify images by simply presenting their requirements in text. In the future, we hope to expand the applications of RDM, such as more flexible control of the position, shape, and size of the generated area.


\bibliographystyle{ACM-Reference-Format}
\bibliography{acmart}


\end{document}